\documentclass{article}

\usepackage[final]{neurips}

\makeatletter
\renewcommand{\@noticestring}{}
\makeatother

\usepackage{float}

\makeatletter
\renewenvironment{table}
  {\setlength{\abovecaptionskip}{\baselineskip}%
   \setlength{\belowcaptionskip}{0pt}%
   \@float{table}}
  {\end@float}
\makeatother

\usepackage[utf8]{inputenc}
\usepackage[T1]{fontenc}
\usepackage{amsmath,amssymb}
\usepackage{amsthm}
\usepackage{graphicx}
\usepackage{booktabs}
\usepackage{xcolor}
\usepackage[hidelinks]{hyperref}
\usepackage{url}

\graphicspath{{figures/}}
\theoremstyle{plain}
\newtheorem{fact}{Fact}

\newcommand{\R}{\mathbb{R}}
\newcommand{\rank}{\operatorname{rank}}
\newcommand{\dmin}{d_{\min}}
\newcommand{\dstate}{d_{\mathrm{state}}}
\newcommand{\recninety}{\mathrm{rec}@0.9}

\title{The Rank the Task Demands:
A Causal Rank Law for Matrix Memories Trained on Group Composition}

\author{%
  Samuel Larson\\
  Pebble ML\\
  \texttt{samlarson@pebbleml.com}%
}

\begin{document}

\maketitle

\begin{abstract}
Matrix-valued memories make rank the natural budget of a learned
representation: the number of independent directions a state spans bounds
what it can bind, compose, and track. We report causal evidence, on a
group-composition testbed trained under a hard single-state bottleneck
with a fixed decoder that cannot launder rank, that gradient descent
recruits precisely the rank the task's algebra demands. A companion
paper \citep{larson2026companion} establishes the analogous recruitment
and causal necessity pattern on a $K$-pair associative-binding testbed,
where exact recovery provably requires state rank at least $K$; this
paper inherits that instrument and extends the rank law from a scalar
capacity bound to a representation-theoretic one.
We train toward chosen minimal faithful reference representations
embedded in larger matrices.
On
group-composition state tracking over five finite groups spanning the
solvable/non-solvable divide, the recruited rank equals the group's minimal
faithful real representation dimension $\dmin$ (Spearman $\rho = 0.9747$,
the design's tie-capped maximum),
the dimension-matched solvable/non-solvable
pair $S_4$/$A_5$ is statistically equivalent under a pre-registered
test,
and a pre-registered force-rank test separates a guaranteed similarity
ceiling from empirical recovery at the target dimension: one rank below
$\dmin$, cosine similarity is capped by the target's
tied unit spectrum at $\sqrt{(\dmin{-}1)/\dmin} \le 0.894$, below the
$0.9$ threshold in every group by construction, with observed cells
at 86--95\% (mean 91\%) of that ceiling;
at $\dmin$, not
guaranteed a priori, recovery clears the pre-registered
anchor-relative bar at four seeds per group in all five groups, both
marquee members included.
Within this testbed, measured effective rank tracks representation
dimension; the matched-dimension $S_4$/$A_5$ comparison establishes
equivalence within the pre-registered tolerance.
\end{abstract}


\section{Introduction}
\label{sec:intro}

A matrix-valued memory represents by spanning: whatever a $d \times d$
state binds or tracks, it holds in the geometry of its column space, and
rank is the budget it spends. Prior work treats this budget
descriptively \citep{nazari2026rank, sun2026staterank} or through
hand-built constructions \citep{nichani2025factual}; neither answers
what a geometric account of learned representations needs answered:
when a task's algebra fixes a minimal representational dimension, does
gradient descent recruit exactly that dimension, and is it causally
load-bearing?

We answer both halves affirmatively under one discipline: a hard
single-state bottleneck (the decoder reads only one matrix $Z$, verified
by a gradient blank-out test) and recovery at a cosine-similarity
threshold of 0.9 (scoring against the true continuous target, never argmax over a codebook,
which would let a rank-1 state recover on the order of $d$ associations
\citep{nichani2025factual}). Section~\ref{sec:setup} includes the
provable $K$-pair-binding foundation. The headline is
group-composition state tracking, where representation theory
supplies the ground truth: across five finite groups spanning the
solvable/non-solvable divide, the minimal faithful real representation
dimension $\dmin$ predicts the recruited rank almost perfectly
(Section~\ref{sec:observed}), a
matched-dimension solvable/non-solvable pair lands statistically
equivalent (Section~\ref{sec:equivalence}),
and a pre-registered force-rank
razor (Section~\ref{sec:razor}) checks a geometrically guaranteed
similarity ceiling and confirms sufficiency in all five groups:
at rank $\dmin{-}1$, the maximum cosine similarity is below the $0.9$ threshold
($\sqrt{(\dmin{-}1)/\dmin} \le 0.894$), and $\dmin$ suffices at four
seeds per group in every group, which is not guaranteed a priori.
Correcting the target's padding (Appendix~\ref{app:instrument}) recovers the
razor's sufficiency result.

\section{Tasks, Models, Instrument, and the Binding Foundation}
\label{sec:setup}

\textbf{Tasks.} In the binding task, each episode presents $K$ key--value
pairs. The encoder writes one state $Z \in \R^{d \times d}$. Prediction is
the literal unbind $Z k_j$. $\recninety$ is the fraction of queries with
cosine to the true value above 0.9. The group task is the word problem of a
finite group $G$. A word $w = g_{i_1} \cdots g_{i_L}$ is drawn as a random
walk on $G$'s Cayley graph ($L \sim \mathrm{U}\{1,\dots,8\}$). It must map
to $\rho_G(g_{i_1} \cdots g_{i_L})$ under a pinned orthogonal reference
representation of dimension $\dmin(G)$, the minimal faithful real
representation dimension. The groups are $S_3, S_4, A_5, S_5, A_6$ with
$\dmin = 2, 3, 3, 4, 5$ (Appendix~\ref{app:groups}). The first two are
solvable, and the rest are not. Non-solvable word problems are
$\mathrm{NC}^1$-complete \citep{barrington1989,barringtontherien1988}.
$\dmin$ is a different, representation-theoretic axis. The design uses
$S_4$/$A_5$, matched at $\dmin = 3$ with opposite solvability, as the
dissociation pair.

We study learning under direct supervision toward a chosen minimal
faithful representation, embedded in a larger matrix. The target
specifies this representation; the rank requirements studied here apply
to this target and scoring setup, rather than to storage across all
possible encodings or decoders.

\textbf{Model.} A Transformer encoder with $\dstate$ learned row-reader
queries maps each word to a single state
$Z(w) \in \R^{\dstate \times \dstate}$, $\dstate = \dmin + 2$, leaving
spare dimensions so over-recruitment is expressible. The target embeds
the reference block-diagonally ($\rho_G(\cdot) \oplus I_2$ in the
observational arm, $\rho_G(\cdot) \oplus 0$ in the causal-razor arm;
Appendix~\ref{app:instrument}). The loss is cosine distance. The fixed
readout has no learned weights that could launder rank, and the decoder
reads only $Z$ (blank-out verified). Convergence bars pinned the per-group step
budgets (8k--40k) before any decisional cell ran, as recorded in the design
document's pre-registration.

\textbf{Instrument.} The learned state is defined only up to scale and an
orthogonal change of basis (a Schur intertwiner):
$Z(w) \approx c\,Q\,[\rho_G(w) \oplus \cdot\,]\,Q^{\top}$. We estimate the
model's own dominant $\dmin$-subspace $U$ from the SVD of the
\emph{centered} covariance of $Z(w)$ over held-out words. We measure
\textbf{restricted effective rank} as the entropy effective rank of
$U^{\top} Z(w) U$. We score \textbf{degauged recovery} as cosine after fitting
$(c, Q)$ on a fitting split and evaluating on a disjoint split (50 fresh
words per cell, 30 to fit and 20 scored, so $\recninety$ moves in steps of
0.05), reported as $\recninety$. Centering is load-bearing (Appendix~\ref{app:instrument}). Rank
is invariant under the fitted gauge, so a rank-deficient state cannot be
degauged into a full-rank one. For force-rank cells, the pre-registration
pinned the conservative full-$Q$ Procrustes variant (\emph{crosscheck}
$\recninety$) as the decisional readout.

\textbf{The provable foundation (binding).}
\begin{fact}
\label{fact:bound}
If $Z k_j = v_j$ exactly for $K$ bindings with linearly independent keys
and values, then $Z K_{\mathrm{mat}} = V_{\mathrm{mat}}$, so
$K = \rank(V_{\mathrm{mat}}) \leq \rank(Z)$.
\end{fact}

The bound is classical \citep{kohonen1972correlation,
anderson1972simple}, holding only under exact continuous recovery. On this
binding testbed, a companion paper \citep{larson2026companion} establishes
that gradient descent recruits effective rank tracking $K$. The same paper
shows that the recruited rank is causally necessary through a train-time
force-rank step at the provable bound. It also shows that the trained operator
composes exactly under repeated self-application. The composition-stability
probe is periodicity-equivalent to depth 5 under the single 8-cycle target
\citep{liu2023shortcuts} (Appendix~\ref{app:period}). This paper inherits that
instrument and carries only the group-composition program.

\section{The Rank Law on Group Composition, Observed}
\label{sec:observed}

\begin{figure}[t]
\centering
\includegraphics[width=\textwidth]{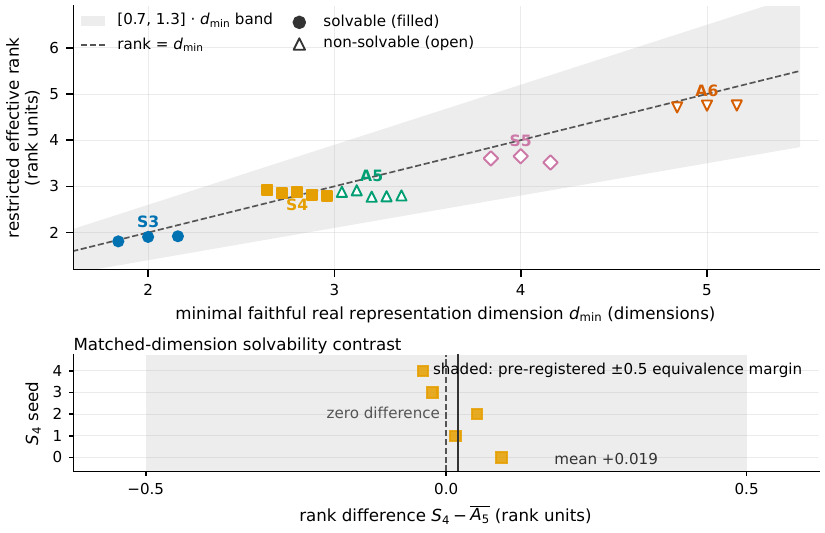}
\caption{Recruited rank follows representation dimension across the five
tested groups. Each point is one seed's restricted effective rank on the
group word problem vs.\ its group's $\dmin$. Filled markers denote solvable
groups; open markers denote non-solvable groups. Small horizontal offsets
separate seeds at the same categorical value. Shading marks the pre-registered
$[0.7, 1.3] \cdot \dmin$ band, and the dashed line marks the identity. All 19
seeds are in band, and $S_4$/$A_5$ coincide at $\dmin = 3$.
The lower panel plots per-seed $S_4 - \overline{A_5}$ rank differences at
matched $\dmin = 3$ against the pre-registered $\pm 0.5$ rank-unit
equivalence margin. Shading marks the margin. The vertical lines mark zero
difference and the observed mean difference, $+0.019$
(Section~\ref{sec:equivalence}).}
\label{fig:tracking}
\end{figure}

Per-group mean restricted effective rank lands within 4.9--10.2\% of $\dmin$
at every group, and all 19 seeds sit inside the pre-registered
$[0.7, 1.3] \cdot \dmin$ band (Figure~\ref{fig:tracking}; means and per-seed
values in Appendix~\ref{app:m1}).

Spearman correlation between recruited rank and $\dmin$ is
$\rho = 0.9747$,
the maximum this design can express, since the $S_4$/$A_5$ tie caps
$\rho$ below 1; under the exact permutation null,
$P(\rho \geq 0.8) = 8/120 \approx 6.7\%$ (0.8 is pre-registered, below the
next achievable level 0.87),
so by pre-registration this leg is corroborating rather than independently
decisive: the band's upper half is non-binding by construction, and a
disclosed length-robustness split is reported alongside it
(Appendix~\ref{app:instrument}).
The dimension-matched pair $S_4$/$A_5$ is tested directly in
Section~\ref{sec:equivalence}.

\section{Equivalence at Matched Dimension}
\label{sec:equivalence}

The design uses $S_4$/$A_5$ as its contrast. The groups share $\dmin = 3$ but
lie on opposite sides of the solvability divide, the axis on which expressivity
theory separates state-tracking architectures
\citep{merrill2024illusion, grazzi2025negative,
siems2025deltaproduct}. This matched-dimension comparison tests whether
measured effective rank is equivalent across the selected pair despite
their opposite solvability. \citet{siems2025deltaproduct} provide the
precursor observation (\S5.2): $S_4$ and $A_5$ both extrapolate with two
Householder reflections and keys of size 3. They attribute this to the groups'
embedding in $\mathrm{SO}(3)$, and their PCA of the trained keys is
three-dimensional. The present test turns that single-architecture reading
into a pre-registered equivalence claim on state rank across the five-group
family. Section~\ref{sec:razor} makes the dimension causal. Because the
interesting outcome is a null, the pre-registration specifies an equivalence
test rather than a difference test. The comparison uses Welch two-one-sided
tests on restricted effective rank at margin $\pm 0.5$ rank-units (half the
spacing of the $\dmin$ ladder), with $n = 5$ seeds per group. The design record
fixed a pre-run power simulation before the sweep ran. It confirms that the
test reliably \emph{rejects} equivalence at a true gap of 1.0 rank-units, so a
real class effect of one ladder step could not have hidden inside the margin.

The observed difference is $+0.019$ rank-units (se 0.037, df 7.8), and
both one-sided tests pass at roughly seven times the critical value
($t = 13.06$ and $14.12$ against $t_{\mathrm{crit}} = 1.865$): the two
one-sided tests declare equivalence.
The two groups' seed clusters are visually coincident in
Figure~\ref{fig:tracking} (lower panel). These results establish
equivalence of effective rank for this matched-dimension pair within the
pre-registered $\pm 0.5$ rank-unit tolerance.

\section{The Causal Razor}
\label{sec:razor}

The decisive test intervenes on rank: the encoder's output is
spectrally truncated to rank $k \in \{\dmin{-}1, \dmin, \dmin{+}1\}$
throughout training, against the zero-padded target
$\rho_G(\cdot) \oplus 0$ of rank exactly $\dmin$, alongside an
unconstrained anchor from the same family. Pre-registered reading:
below $\dmin$, a sound readout pins recovery to zero by geometry,
making that arm an integrity control whose registered trigger (any
recovery there) would indicate an instrument leak; the live prediction
is that $k \geq \dmin$ recovers past $0.9\times$ the anchor's
crosscheck $\recninety$.

\begin{figure}[t]
\centering
\includegraphics[width=\textwidth]{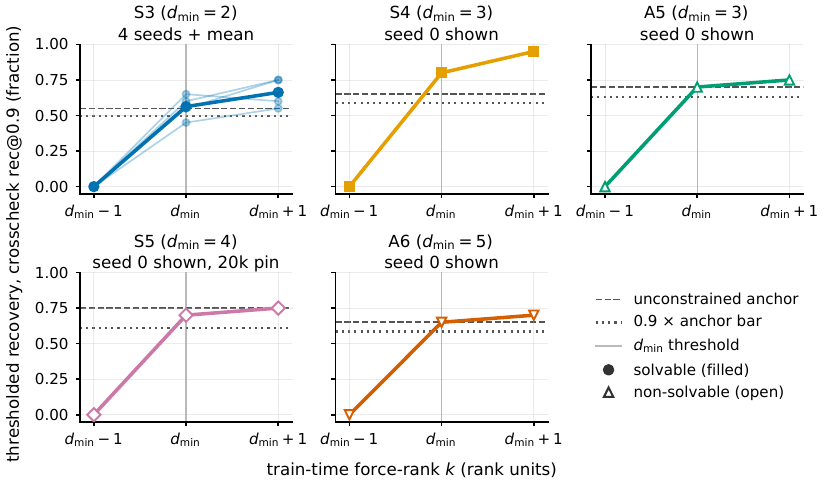}
\caption{Recovery at a cosine-similarity threshold of 0.9 is a step function at $\dmin$: per group,
crosscheck $\recninety$ on held-out words at force-rank
$k \in \{\dmin{-}1, \dmin, \dmin{+}1\}$, vs.\ the unconstrained anchor
(dashed) and the $0.9\times$anchor bar (dotted).
$S_3$ overlays its four seeds (thin lines; bold mean); its
below-$\dmin$ zero is unanimous. The other four groups are drawn at
seed 0; their four-seed sufficiency verdicts are in
Table~\ref{tab:razor}. $S_5$'s dotted bar is its decisional bar,
$0.9\times$ its four-seed anchor mean at its 20k pin
(Appendix~\ref{app:ext4causal}).}
\label{fig:razor}
\end{figure}

\begin{table}[t]
\centering
\footnotesize
\setlength{\tabcolsep}{5pt}
\begin{tabular}{lcccccc}
\toprule
$G$ & anchor & $k{=}\dmin{-}1$ & $k{=}\dmin$ & $k{=}\dmin{+}1$ & bar & $k{=}\dmin$, seed-mean ($n{=}4$) \\
\midrule
$S_3$ & 0.550 & 0.000 & 0.450$^{\ast}$ & 0.550 & 0.495 & \textbf{0.5625} \\
$S_4$ & 0.650 & 0.000 & \textbf{0.800} & 0.950 & 0.585 & \textbf{0.6875} \\
$A_5$ & 0.700 & 0.000 & \textbf{0.700} & 0.750 & 0.630 & \textbf{0.6500} \\
$S_5$ & 0.750 & 0.000 & \textbf{0.700} & 0.750 & 0.6075$^{\dagger}$ & \textbf{0.7000} \\
$A_6$ & 0.650 & 0.000 & \textbf{0.650} & 0.700 & 0.585 & \textbf{0.6750} \\
\bottomrule
\end{tabular}
\caption{The causal razor on the zero-padded rank-$\dmin$ target
(crosscheck $\recninety$ by force-rank $k$). The anchor and $k$
columns are seed 0, one seed per cell; the bar is $0.9\times$ the seed-0
anchor, fixed before any further seed ran ($S_5$: see $\dagger$), and
bold there marks a seed-0 cell that clears it. The last column is the
seed-mean of $k{=}\dmin$ over seeds 0--3, judged against that bar by
the pre-registered rule: all five groups confirm (bold). Exactly 0.000
one rank below $\dmin$ in all four seeds of every group (the forced
ceiling is $\le 0.894$). Per-seed own-bar clears ($0.9\times$ each
seed's own anchor; disclosed, not decisional): $S_3$ 2/4, $S_4$ 4/4,
$A_5$ 4/4, $S_5$ 4/4, $A_6$ 4/4.
$^{\ast}$$S_3$'s seed-0 cell sat inside the pre-stated $\pm 0.05$
marginality trigger and was extended first (seed-mean 0.5625 against
the fixed 0.495 bar); the other four groups were extended under a
separate pre-registration (Appendices~\ref{app:s3causal},
\ref{app:ext4causal}).
$^{\dagger}$$S_5$'s step pin is 20k, set by a separately
pre-registered re-test (Appendix~\ref{app:ext4causal}); by that
pre-registration its bar is $0.9\times$ its four-seed anchor mean
(0.6750), since no seed-0 literal existed at that pin.}
\label{tab:razor}
\end{table}

Table~\ref{tab:razor} and Figure~\ref{fig:razor} show the result on the
pre-pinned crosscheck readout. Necessity below $\dmin$ is an analytically
forced floor: against a $\dmin$-tied-unit-singular-value target, the
maximum cosine any rank-$(\dmin{-}1)$ operator can reach is
$\sqrt{(\dmin{-}1)/\dmin} \le 0.894$, below the $0.9$ threshold in every
group by construction, so $\recninety = 0.000$ at $k = \dmin{-}1$ follows
regardless of training quality; observed cells still reach 86--95\% of
that ceiling (mean 91\%; Appendix~\ref{app:instrument}), evidence of
near-optimal training under the cap.
The zero is noiseless in all four seeds of every group.
Sufficiency at $\dmin$ carries the causal claim, since no such bound
guarantees a rank-exactly-$\dmin$ solution is reachable. Every group
was run at four seeds, and the pre-registered rule compares the
seed-mean of $k{=}\dmin$ against the bar fixed from seed 0 before the
further seeds ran, never against a bar recomputed from the seeds being
judged. $S_3$, whose seed-0 cell sat inside the pre-stated $\pm 0.05$
marginality trigger, confirms at seed-mean 0.5625 against the fixed
0.495 bar;
$S_4$, $A_5$, and $A_6$ confirm at seed-means 0.6875, 0.6500, and
0.6750 against fixed bars 0.585, 0.630, and 0.585, each clearing its
own-seed bar in 4 of 4 seeds; $S_5$ confirms at seed-mean 0.7000
against its bar 0.6075, per-seed $k{=}\dmin$ 0.700/0.600/0.750/0.750
against anchors 0.750/0.600/0.700/0.650, own-bar clears 4 of 4, and
$k{=}\dmin{-}1$ at 0.000 in all four seeds; its bar is $0.9\times$ its
four-seed anchor mean rather than a seed-0 literal because its 20k
step pin was set by a separately pre-registered re-test
(Appendix~\ref{app:ext4causal}). Sufficiency therefore holds at four
seeds in all five groups (per-seed detail in
Appendices~\ref{app:s3causal} and \ref{app:ext4causal}).
Both marquee groups confirm at $\dmin$, meeting the causal criterion.
Table~\ref{tab:wmrank} adds an instrument-independent reading: on the
zero-padded target the unconstrained anchor's whole-matrix effective
rank, taken over the full $\dstate \times \dstate$ state with no
subspace lens, lands at $1.82/2.82/2.83/3.77/4.67$ (four-seed means)
against $\dmin = 2/3/3/4/5$ with two ambient dimensions available, so
the recruited rank equals $\dmin$ without any restriction to the
model's dominant subspace.

Identity-padding the target confounds the razor by handing every capped
arm a free rank-2 loss reduction (Appendix~\ref{app:instrument});
the zero-padded target above removes the confound.

\section{Related Work and Limitations}
\label{sec:related}

\citet{chughtai2023universality} reverse-engineers group multiplication via
representation theory. \citet{stander2024cosets} and \citet{wu2025unified}
dispute and refine that account (coset circuits; verified per-argument
equivariance). \citet{he2026spectral} prove that two-layer networks on
$g_1 \star g_2$ learn spectral representations with a rank-one alignment of
Fourier coefficients. \citet{shutman2025words} frame word learning as a
low-rank 3-tensor. All four study the binary operation, not a sequential state,
and none relates a learned dimension to $\dmin$. On sequential composition,
\citet{marchetti2026sequential} show recurrent networks acquire irreps one at a
time with width scaling in the irrep dimensions on cyclic and dihedral groups.
\citet{zhang2026recurrence} find on looped transformers that
$\mathrm{NC}^1$-completeness costs nothing while group order does. Neither
measures state rank. \citet{kazanskii2026grokking} regularizes
representation dimensionality on permutation composition to speed
grokking, without a necessity threshold. Here, rank itself, not an irrep
decomposition, is the measured and manipulated quantity, and the
required dimension is fixed in closed form by $\dmin$.
\citet{nazari2026rank} and \citet{sun2026staterank} measure
state-rank dynamics in pretrained linear-attention models observationally,
with no task algebra fixing a required dimension. They read low rank as
under-used capacity, whereas this paper reads it as the rank the task demands.
\citet{parnichkun2025ess} apply an entropy effective rank to sequence-model
memory as a utilization measure. \citet{truong2026spectral} track
spectral-entropy collapse of representation covariances on group tasks. The
estimator here is therefore standard, not new.
\citet{mishra2026m2rnn} train matrix-state RNNs on $S_3$ composition
without a rank intervention. \citet{grazzi2025negative},
\citet{siems2025deltaproduct}, \citet{merrill2024illusion},
\citet{shakerinava2026diagonal}, and \citet{nowak2026algebraic}
characterize which word problems recurrent architectures can express. They
sort the problems by solvability and by the depth a subnormal series demands,
an axis orthogonal to the one measured here (dimension governs state rank;
series length governs depth). \citet{deletang2023chomsky} benchmark sequence
architectures across the formal-language hierarchy. The equivalence test
instead sorts by representation dimension. A bolt-on-latent negative result
\citep{larson2026gradient} is this result's counterpart.
\citet{nichani2025factual} and \citet{barnfield2026sharp} prove
associative-memory capacity under argmax (winner-take-all) decoding,
whereas we score cosine similarity against continuous targets and report
recovery at a threshold of 0.9.
\citet{huh2024platonic} propose that models trained on different tasks
converge to a shared representation; the present law is a narrower,
algebra-driven instance of that convergence.

\textbf{Limitations.} All results are on sub-1M-parameter synthetic
testbeds, not claims about pretrained language models. Razor cells are
four seeds per group; $S_5$'s step pin is 20k rather than 8k, set by a
pre-registered re-test after the 8k pin failed convergence
(Appendix~\ref{app:ext4causal}).
$S_3$ carries a soft-convergence flag at its own shortest pinned budget
(Appendix~\ref{app:gate1a}), so its confirmation reads directionally.
A companion paper \citep{larson2026companion}
carries the binding-task program (the recruitment grid, force-rank
staircase, and composition mechanism) in full; this paper shares no
figures or tables with it.

\bibliographystyle{plainnat}
\bibliography{refs}

\appendix

\section{The Five Groups and Reference Representations}
\label{app:groups}

\begin{table}[H]
\centering
\small
\begin{tabular}{lcccl}
\toprule
$G$ & $|G|$ & $\dmin$ & solvable & realization of $\rho_G$ \\
\midrule
$S_3$ & 6   & 2 & yes & standard (dihedral) \\
$S_4$ & 24  & 3 & yes & cube rotations \\
$A_5$ & 60  & 3 & no  & icosahedral rotations \\
$S_5$ & 120 & 4 & no  & standard (zero-sum) \\
$A_6$ & 360 & 5 & no  & standard (zero-sum) \\
\bottomrule
\end{tabular}
\caption{The five task groups. $\dmin$ is the minimal faithful real
representation dimension; each reference representation is orthogonal by
construction and verified faithful (injective on $G$) at build time.
$S_4$/$A_5$ form the marquee pair: matched $\dmin$, opposite solvability.}
\label{tab:groups}
\end{table}

\section{Per-Seed Observational Rank}
\label{app:m1}

\begin{table}[H]
\centering
\small
\begin{tabular}{lccccc}
\toprule
$G$ & $\dmin$ & $n$ & restricted eff.\ rank & band & in band \\
\midrule
$S_3$ & 2 & 3 & $1.877 \pm 0.060$ & $[1.4, 2.6]$ & 3/3 \\
$S_4$ & 3 & 5 & $2.852 \pm 0.054$ & $[2.1, 3.9]$ & 5/5 \\
$A_5$ & 3 & 5 & $2.832 \pm 0.062$ & $[2.1, 3.9]$ & 5/5 \\
$S_5$ & 4 & 3 & $3.591 \pm 0.069$ & $[2.8, 5.2]$ & 3/3 \\
$A_6$ & 5 & 3 & $4.736 \pm 0.023$ & $[3.5, 6.5]$ & 3/3 \\
\bottomrule
\end{tabular}
\caption{Restricted effective rank of the unconstrained trained state
(mean $\pm$ sd over seeds) against each group's $\dmin$, with the
pre-registered $[0.7, 1.3] \cdot \dmin$ band. Every seed of every group is
in band; Spearman $\rho = 0.9747$ is the tie-capped maximum; the
dimension-matched pair $S_4$/$A_5$ differs by 0.019 rank-units.}
\label{tab:m1}
\end{table}

Per-group deviation of the mean from $\dmin$ (Table~\ref{tab:m1}):
$S_3$ 6.1\%, $S_4$ 4.9\%, $A_5$ 5.6\%, $S_5$ 10.2\%, $A_6$ 5.3\%.

$S_4$'s seed-0 razor $k{=}\dmin$ cell (Table~\ref{tab:razor}) numerically exceeds its own unconstrained anchor by 0.15;
plausibly the capped arm skips the ambient dimensions the zero-padded
anchor must still learn to null under the same fixed step budget, an
unconfirmed hypothesis.
At four seeds the inversion holds strictly in 2 of 4 $S_4$ seeds and
ties in the other 2 (Appendix~\ref{app:ext4causal}), so it stays
unconfirmed.

\section{Whole-Matrix Effective Rank}
\label{app:wmrank}

\begin{table}[H]
\centering
\small
\begin{tabular}{lcccc}
\toprule
$G$ & $\dmin$ & $\dstate$ & observational arm ($\rho_G \oplus I_2$) & zero-padded anchor ($\rho_G \oplus 0$) \\
\midrule
$S_3$ & 2 & 4 & $3.769 \pm 0.071$ ($n{=}3$) & $1.824 \pm 0.064$ ($n{=}4$) \\
$S_4$ & 3 & 5 & $4.835 \pm 0.066$ ($n{=}5$) & $2.817 \pm 0.101$ ($n{=}4$) \\
$A_5$ & 3 & 5 & $4.781 \pm 0.066$ ($n{=}5$) & $2.833 \pm 0.077$ ($n{=}4$) \\
$S_5$ & 4 & 6 & $5.510 \pm 0.084$ ($n{=}3$) & $3.767$ (3.705--3.860, $n{=}4$) \\
$A_6$ & 5 & 7 & $6.717 \pm 0.022$ ($n{=}3$) & $4.669 \pm 0.097$ ($n{=}4$) \\
\bottomrule
\end{tabular}
\caption{Whole-matrix effective rank of the unconstrained trained state:
the entropy effective rank of the full $\dstate \times \dstate$ state
with no subspace lens, so unlike the restricted quantity of
Table~\ref{tab:m1} it can range up to $\dstate = \dmin + 2$. Observational
arm: mean $\pm$ sd over seeds against the eye-padded target of rank
$\dmin + 2$, which the state recruits ($3.77/4.84/4.78/5.51/6.72$ against
$\dstate = 4/5/5/6/7$). Zero-padded anchor: the
causal wave's unconstrained arm against the rank-$\dmin$ target, mean
$\pm$ sd over four seeds in every group ($S_5$ at its 20k pin,
Appendix~\ref{app:ext4causal}, reported as mean and range). On the
zero-padded target the unconstrained state recruits $\dmin$
($1.82/2.82/2.83/3.77/4.67$ against $\dmin = 2/3/3/4/5$) and leaves the
two spare dimensions unused.}
\label{tab:wmrank}
\end{table}

\begin{table}[H]
\centering
\small
\begin{tabular}{lcccc}
\toprule
$G$ & $k{=}\dmin{-}1$ & $k{=}\dmin$ & $k{=}\dmin{+}1$ & unconstrained \\
\midrule
$S_3$ & $1.000 \pm 0.000$ & $1.816 \pm 0.081$ & $1.841 \pm 0.053$ & $1.824 \pm 0.064$ \\
$S_4$ & 1.953 & 2.937 & 2.964 & 2.952 \\
$A_5$ & 1.991 & 2.824 & 2.869 & 2.882 \\
$S_5$ & 2.931 & 3.766 & 3.801 & 3.860 \\
$A_6$ & 3.887 & 4.631 & 4.708 & 4.730 \\
\bottomrule
\end{tabular}
\caption{Whole-matrix effective rank of every zero-padded razor cell
(seed 0; $S_3$ mean $\pm$ sd over its four seeds; $S_5$ at its 20k
pin), for completeness alongside Table~\ref{tab:wmrank}. The capped cells sit at or below their
cap $k$; at $k \geq \dmin$ they land where the unconstrained anchor does.}
\label{tab:wmrank_cells}
\end{table}

\section{Instrument Details and the Two Defects}
\label{app:instrument}

\textbf{Centering.} For an orthogonal target the uncentered covariance
of $Z(w)$ is isotropic and carries no subspace information; the
production lens centers it (a nontrivial irreducible representation has
zero group mean, cancelling the constant block). The step is
load-bearing: an uncentered lens scores a flawless synthetic model 0.705
where the centered lens scores 0.9996 on identical data.

\textbf{Length-robustness split.} The disclosed split referenced in
Section~\ref{sec:observed} scores only words of length $L \geq 2$
(the $L = 1$ read is attention-degenerate and was demoted with a
mechanism note in the run records): it moves per-group mean recovery
cosine by at most 0.013 (per-seed at most 0.041), with no
group-selective divergence.

\textbf{The ambient-identity tax.} The first 58-cell sweep's force-rank
target was the eye-padded $\rho_G(\cdot) \oplus I_2$, whose rank is
$\dstate = \dmin + 2$ for every group.
Because the identity block is constant across words, it is the
cheapest reliable loss reduction available, and a rank-capped arm buys it
before the group representation; the residual budget for $\rho_G$ is
$k - 2$, below $\dmin$ at every grid point, so no capped cell under
identity-padding ever tested the law's confirm direction. Three raw-artifact
signatures established this: (i) force-rank cells' direct cosine matched
the rank-$k$ optimum $\sqrt{k/\dstate}$ with mean absolute deviation 0.028
across all 39 cells (maximum 0.166; the two $S_5$ below-$\dmin$ cells are
the only outliers, an additional disclosed optimization shortfall), so the
arms trained to their rank-constrained ceiling rather than under-training;
(ii) centered restricted rank of capped arms landed near $k - 2$, the
residual budget; (iii) in the zero-padded grid, a deliberately
eye-padded corroboration arm reproduces the failure on demand at raw
$k = \dmin{+}1$ (effective budget $\dmin{-}1$) while the tax-paid point
recovers.
The sweep verdict was registered as inconclusive with the tax as
mechanism; the zero-padded causal wave of Section~\ref{sec:razor} is
the registered fix.
The zero-padded grid's 30 cells were configuration-verified one-by-one
against a manifest re-derived independently from the design record (steps,
padding mode, and force-rank value per cell; zero skipped steps).

\textbf{The observational band's non-binding upper half.}
\texttt{entity\_subspace\_from\_words} returns the top-$\dmin$ left
singular vectors of the centered covariance by construction, so the
restricted state $U^{\top} Z U$ is always $\dmin \times \dmin$, and the
entropy effective rank of a $\dmin \times \dmin$ input has range
$[1, \dmin]$. Restricted effective rank therefore cannot exceed $\dmin$,
and so cannot exceed $1.3 \cdot \dmin$, for any seed of any group; only
the band's lower half, $\geq 0.7 \cdot \dmin$, is a test the data could
fail. Every group mean lands in $[0.898, 0.951] \cdot \dmin$.

\textbf{The rank-constrained cosine ceiling.} The causal wave's target
is $\rho_G(\cdot) \oplus 0$, of rank exactly $\dmin$ with all $\dmin$
informative singular values equal to 1 (the reference representation is
orthogonal by construction, Appendix~\ref{app:groups}). For any matrix
$M$ of rank $\leq k \leq \dmin$, von Neumann's trace inequality gives
$\langle M, T\rangle_F \leq \sum_{i=1}^{k} \sigma_i(M)\,\sigma_i(T) =
\sum_{i=1}^{k} \sigma_i(M)$, and Cauchy--Schwarz bounds
$\sum_{i=1}^{k} \sigma_i(M) \leq \sqrt{k}\,\lVert M \rVert_F$, so
$\cos(M, T) \leq \sqrt{k/\dmin}$. At $k = \dmin{-}1$ this ceiling is
$0.707/0.816/0.816/0.866/0.894$ for $S_3/S_4/A_5/S_5/A_6$, below the
$\recninety$ threshold of 0.9 in every case: no rank-$(\dmin{-}1)$
state, however trained, can register a single word above cosine 0.9
against this target. The necessity result ($\recninety = 0.000$ at
$\dmin{-}1$, all groups, all seeds) is this bound realized, not an
independent discovery about what SGD finds; the corresponding empirical
content is that the observed cosines at $\dmin{-}1$ (0.61--0.84, seed 0) sit
under their respective rank-constrained optima, evidence optimization
reaches the geometry's own ceiling. At $k \geq \dmin$ the ceiling is
1.0 (exact recovery is representable), so recovery above 0 at $\dmin$
is not guaranteed by the metric and carries the causal weight
(sufficiency).
As a fraction of the geometrically forced ceiling
$\sqrt{(\dmin{-}1)/\dmin}$, the below-$\dmin$ crosscheck cosines read
$S_3$ 86.3\%, $S_4$ 91.2\%, $A_5$ 94.9\%, $S_5$ 87.9\%, and $A_6$
93.5\% (mean 90.8\%; seed-0 cells, $S_5$ at its 20k pin).

\textbf{Estimator dependence.} The pre-registered instrument is the
entropy effective rank; under the stable-rank estimator (squared
Frobenius norm over squared spectral norm, the estimator of
\citet{nazari2026rank}) the same unconstrained states read
$1.451/2.156/2.078/2.362/3.340$ against $\dmin = 2/3/3/4/5$, that is
$0.59$--$0.73 \cdot \dmin$, outside the pre-registered band, which was
written for the entropy estimator, while the group ordering is
unchanged (Spearman 0.975 on group means). The ordinal law is therefore
estimator-robust; the point estimate ``equals $\dmin$'' is a property
of the entropy estimator on a near-flat spectrum with one weak
direction, which stable rank penalizes; and the causal razor caps
algebraic rank and is estimator-free.

\textbf{Primary and crosscheck degauging.} The design record's default
pipeline treats scale-only degauging ($\hat{Q} = I$) as the primary
metric, with the fitted-$(c, Q)$ Procrustes score retained as a
robustness cross-check, after calibration on unconstrained checkpoints
found $\hat{Q} \approx I$ empirically. Under the force-rank grid's
zero-padded target this equivalence breaks: the informative block's
degenerate singular spectrum makes the scale-only fit's basis
arbitrary, so the pre-registration for the causal wave specifically
pins the fitted-$(c, Q)$ score, denoted crosscheck $\recninety$
throughout Section~\ref{sec:razor} and Table~\ref{tab:razor} and Figure~\ref{fig:razor}, as decisional; the scale-only score informs no
reported conclusion. The two diverge sharply on this grid, which is why
the distinction is disclosed here.

\section{$S_3$ Per-Seed Causal Detail}
\label{app:s3causal}

\begin{table}[H]
\centering
\small
\begin{tabular}{lcccccc}
\toprule
seed & anchor & $k{=}\dmin{-}1$ & $k{=}\dmin$ & $k{=}\dmin{+}1$ & own bar & clears \\
\midrule
0 & 0.550 & 0.000 & 0.450 & 0.550 & 0.495 & no ($-0.045$) \\
1 & 0.600 & 0.000 & 0.550 & 0.750 & 0.540 & yes ($+0.010$) \\
2 & 0.800 & 0.000 & 0.600 & 0.750 & 0.720 & no ($-0.120$) \\
3 & 0.600 & 0.000 & 0.650 & 0.600 & 0.540 & yes ($+0.110$) \\
\bottomrule
\end{tabular}
\caption{$S_3$ four-seed causal detail (crosscheck $\recninety$) behind
the seed-mean confirmation in Section~\ref{sec:razor}; the own bar is
$0.9 \times$ each seed's anchor. $k = \dmin{-}1$ reads exactly 0.000 in
all four seeds; $k = \dmin$ clears each seed's own bar in 2 of 4 (seeds
1, 3), missing in the other 2 because the anchor itself ranges
0.550--0.800 across seeds. The pre-registered criterion compares the
seed-mean $k = \dmin$ value (0.5625) against the bar fixed before the
extension ran (0.495, from seed 0's anchor), specifically to avoid
computing the bar from the same noisy seeds it is judged against;
recomputing the bar as $0.9\times$ the four-seed anchor mean (raw mean
0.6375, bar 0.574) would put the seed-mean value 0.011 below it.}
\label{tab:s3seeds}
\end{table}

\begingroup
\setlength{\intextsep}{0pt}
\section{Per-Seed Razor Detail for $S_4$, $A_5$, $S_5$, $A_6$}
\label{app:ext4causal}

\begin{table}[H]
\centering
\footnotesize
\setlength{\tabcolsep}{4pt}
\begin{tabular}{lccccccc}
\toprule
seed & anchor & $k{=}\dmin{-}1$ & $k{=}\dmin$ & $k{=}\dmin{+}1$ & own bar & clears & gate1a (anchor / $k{=}\dmin$) \\
\midrule
\multicolumn{8}{l}{$S_4$ ($\dmin = 3$, 20k steps): seed-mean 0.6875 vs.\ fixed bar 0.585, confirmed; self-referential bar 0.5625} \\
0 & 0.650 & 0.000 & 0.800 & 0.950 & 0.585 & yes ($+0.215$) & --- \\
1 & 0.500 & 0.000 & 0.600 & 0.600 & 0.450 & yes ($+0.150$) & --- \\
2 & 0.700 & 0.000 & 0.700 & 0.650 & 0.630 & yes ($+0.070$) & --- \\
3 & 0.650 & 0.000 & 0.650 & 0.650 & 0.585 & yes ($+0.065$) & --- \\
\midrule
\multicolumn{8}{l}{$A_5$ ($\dmin = 3$, 20k steps): seed-mean 0.6500 vs.\ fixed bar 0.630, confirmed; self-referential bar 0.5625} \\
0 & 0.700 & 0.000 & 0.700 & 0.750 & 0.630 & yes ($+0.070$) & --- \\
1 & 0.700 & 0.000 & 0.750 & 0.750 & 0.630 & yes ($+0.120$) & --- \\
2 & 0.550 & 0.000 & 0.600 & 0.550 & 0.495 & yes ($+0.105$) & --- \\
3 & 0.550 & 0.000 & 0.550 & 0.450 & 0.495 & yes ($+0.055$) & --- \\
\midrule
\multicolumn{8}{l}{$S_5$ ($\dmin = 4$, 20k steps, pre-registered re-test): seed-mean 0.7000 vs.\ bar 0.6075, confirmed} \\
0 & 0.750 & 0.000 & 0.700 & 0.750 & 0.675 & yes ($+0.025$) & 0.981 / 0.981 \\
1 & 0.600 & 0.000 & 0.600 & 0.650 & 0.540 & yes ($+0.060$) & 0.974 / 0.960 \\
2 & 0.700 & 0.000 & 0.750 & 0.650 & 0.630 & yes ($+0.120$) & 0.974 / 0.970 \\
3 & 0.650 & 0.000 & 0.750 & 0.700 & 0.585 & yes ($+0.165$) & 0.959 / 0.967 \\
\midrule
\multicolumn{8}{l}{$A_6$ ($\dmin = 5$, 40k steps): seed-mean 0.6750 vs.\ fixed bar 0.585, confirmed; self-referential bar 0.5850} \\
0 & 0.650 & 0.000 & 0.650 & 0.700 & 0.585 & yes ($+0.065$) & --- \\
1 & 0.600 & 0.000 & 0.650 & 0.700 & 0.540 & yes ($+0.110$) & --- \\
2 & 0.850 & 0.000 & 0.850 & 0.800 & 0.765 & yes ($+0.085$) & --- \\
3 & 0.500 & 0.000 & 0.550 & 0.550 & 0.450 & yes ($+0.100$) & --- \\
\bottomrule
\end{tabular}
\caption{Four-seed causal detail (crosscheck $\recninety$) for the four
groups extended under a pre-registration recorded before launch, on the
same build, manifest, and decisional readout as seed 0; $S_4$, $A_5$,
and $A_6$ at their seed-0 step pins, $S_5$ at a 20k pin. The own bar is
$0.9\times$ each seed's anchor and is disclosed, not decisional.
$k = \dmin{-}1$ reads exactly 0.000 in all 16 cells. The decisional
comparison is the seed-mean of $k = \dmin$ over seeds 0--3 against the
bar fixed before the seeds ran: from seed 0 for $S_4$, $A_5$, $A_6$
(0.585/0.630/0.585), and, for $S_5$, $0.9\times$ the seed-mean of its
four 20k anchors (0.6075), since no seed-0 literal existed at that pin
and the rule was fixed before launch. $S_4$ 0.6875, $A_5$ 0.6500,
$S_5$ 0.7000, and $A_6$ 0.6750 confirm. The self-referential bar
($0.9\times$ the four-seed anchor mean: 0.5625/0.5625/0.5850 for
$S_4$/$A_5$/$A_6$) is reported for symmetry with
Table~\ref{tab:s3seeds}. Own-bar clears: 4/4 in every group. gate1a
(min validation cosine over $L \in [2, 5]$, bar 0.92) is tabulated for
$S_5$, whose pin was set on that criterion.
$S_5$'s original 8,000-step pin under-trained it (every 8k razor cell
failed the 0.92 convergence bar, Appendix~\ref{app:gate1a}, and the
four-seed $k{=}\dmin$ mean at that pin, 0.4125, fell short of the
fixed 0.450 bar); the pin
was raised to 20,000 steps under a pre-registration recorded before
the re-run, and the 8k cells remain archived with the design record.}
\label{tab:ext4seeds}
\end{table}

\section{Soft Convergence at the Shortest Pinned Budgets}
\label{app:gate1a}

\begin{table}[H]
\centering
\small
\begin{tabular}{lccccc}
\toprule
$G$ & anchor & $k{=}\dmin{-}1$ & $k{=}\dmin$ & $k{=}\dmin{+}1$ & bar (0.92) \\
\midrule
$S_3$ & 0.914 & 0.665 & 0.900 & 0.903 & fail (all 4) \\
$S_5$ & 0.876 & 0.801 & 0.879 & 0.877 & fail (all 4) \\
\bottomrule
\end{tabular}
\caption{Convergence health (gate1a, min validation cosine over
$L \in [2, 5]$) for every seed-0 $S_3$ razor cell at its shortest pinned
step budget and every seed-0 $S_5$ razor cell at its original 8k budget. All
eight
cells fall short of the 0.92 margin; $S_5$'s 0.876--0.879 range is a
larger shortfall than $S_3$'s 0.900--0.914. The necessity leg
(Section~\ref{sec:razor}) is unaffected by construction; $S_3$'s
sufficiency leg should be read as directionally consistent under
disclosed soft convergence. At $S_5$'s 20k pin
(Appendix~\ref{app:ext4causal}), every $S_5$ anchor and $k{=}\dmin$
cell passes gate1a (0.959--0.981), which is why that pin moved.}
\label{tab:gate1a}
\end{table}

\section{Depth-21 Periodicity Under the Single $K$-Cycle}
\label{app:period}
\enlargethispage{2\baselineskip}

The binding task's held-out composition probe uses a single Hamiltonian
$K$-cycle, under which nominal depth 21 shares its target with the
already-tested depth 5 ($\pi^{21} = \pi^{21 \bmod 8} = \pi^5$ exactly).
Appendix A of a companion paper \citep{larson2026companion} performs
this periodicity analysis and the associated depth-decay measurement in
full on the binding testbed; see that appendix for the derivation and
the per-hop predicted-vs-measured table.
\endgroup

\end{document}